\documentclass[preprint,12pt]{elsarticle}

\usepackage{graphics}
\usepackage{amsmath}
\usepackage{amssymb}
\usepackage{amsthm}

\usepackage{algorithm}
\usepackage{algorithmic}

\usepackage{multirow}

\usepackage{pgfplots}
\newcommand{\deff}{\stackrel{\text{def}}{=}}
\newcommand{\abs}[1]{\left|#1\right|}
\newcommand{\norm}[1]{\left\|#1\right\|}
\newcommand{\set}[2]{\left\{ #1 \, \left| \, #2 \right. \right\}}
\newcommand{\block}[2]{{#1}^{(#2)}}
\newcommand{\argmin}[1]{\underset{#1}{\arg\min}\,}

\newcommand{\itemdef}[1]{#1.}
\newcommand{\itemref}[1]{(#1)}

\DeclareMathOperator*{\sgn}{sgn}
\DeclareMathOperator*{\diag}{diag}

\DeclareMathOperator*{\E}{E}
\DeclareMathOperator*{\Err}{Err}
\DeclareMathOperator*{\err}{err}

\DeclareMathOperator*{\Tpr}{tpr}
\DeclareMathOperator*{\Ppv}{ppv}
\DeclareMathOperator*{\minimize}{minimize}

\newtheorem{thm_def}{Definition}
\newtheorem{thm_lemma}{Lemma}
\newtheorem{thm_cor}{Corollary}

\newtheorem{thm_prop}{Proposition}
\newtheorem{thm_rem}{Remark}

\journal{Computational Statistics \& Data Analysis}

\begin{document}

\begin{frontmatter}

\title{Sparse group lasso and high dimensional multinomial
classification.}

\author{Martin Vincent \footnote{Corresponding author. Tel.: +4522860740 \\
E-mail address: vincent@math.ku.dk (M. Vincent).}, Niels Richard Hansen}

\address{University of Copenhagen,
Department of Mathematical Sciences, 
Universitetsparken 5,
2100 Copenhagen \O, Denmark}

\begin{abstract}
The sparse group lasso optimization problem is solved using a coordinate gradient descent
algorithm. The algorithm is applicable to a broad class of convex loss functions.
Convergence of the algorithm is established, and the algorithm is used to investigate the
performance of the multinomial sparse group lasso classifier. On three different real data
examples the multinomial group lasso clearly outperforms multinomial lasso in terms of
achieved classification error rate and in terms of including fewer features for the
classification. The run-time of our sparse group lasso implementation is of the same order of
magnitude as the multinomial lasso algorithm implemented in the R package glmnet. Our
implementation scales well with the problem size. One of the high dimensional examples
considered is a 50 class classification problem with 10k features, which amounts to
estimating 500k parameters. The implementation is available as the R package msgl.
\end{abstract}


\end{frontmatter}

\section{Introduction}

The sparse group lasso is a regularization method that combines the lasso \cite{Tibshirani94regressionshrinkage} and the group
lasso \cite{Meier08thegroup}. Friedman et al. \cite{Friedman:1232025} proposed a coordinate descent approach for the sparse group
lasso optimization problem. Simon et al. \cite{NoahSimonSgl2012} used a generalized gradient descent algorithm
for the sparse group lasso and considered applications of this method for linear, logistic and
Cox regression. We present a sparse group lasso algorithm suitable for high dimensional
problems. This algorithm is applicable to a broad class of convex loss functions. In the
algorithm we combine three non-differentiable optimization methods: the coordinate gradient
descent \cite{tseng2009coordinate}, the block coordinate descent  \cite{tseng2001convergence} and a modified coordinate descent method.

Our main application is to multinomial regression. In Section \ref{sec:sgl} we introduce the general
sparse group lasso optimization problem with a convex loss function. Part \ref{sec:multi} investigates the
performance of the multinomial sparse group lasso classifier. In Part \ref{sec:algo} we present the
general sparse group lasso algorithm and establish convergence.

The formulation of an efficient and robust sparse group lasso algorithm is not straight
forward due to non-differentiability of the penalty. Firstly, the sparse group lasso penalty is
not completely separable, which is problematic when using a standard coordinate descent
scheme. To obtain a robust algorithm an adjustment is necessary. Our solution is a minor
modification of the coordinate descent algorithm; it efficiently treats the singularity at zero
that cannot be separated out. Secondly, our algorithm is a Newton type algorithm; hence we
sequentially optimize penalized quadratic approximations of the loss function. This approach
raises another challenge: how to reduce the costs of computing the Hessian? In Section \ref{sec:ubound}
we show that an upper bound on the Hessian is sufficient to determine whether the minimum
over a block of coefficients is attained at zero. This approach enables us to update a large
percentage of the blocks without computing the complete Hessian. In this way we reduce the
run-time, provided that the upper bound of the Hessian can be computed efficiently. We
found that this approach reduces the run-time on large data sets by a factor of more than 2.

Our focus is on applications of the multinomial sparse group lasso to problems with many
classes. For this purpose we choose three multiclass classification problems. We found that
the multinomial group lasso and sparse group lasso perform well on these problems. The
error rates were substantially lower than the best obtained with multinomial lasso, and the
low error rates were achieved for models with fewer features having non-zero coefficients.
For example, we consider a text classification problem consisting of Amazon reviews with 50
classes and 10k textual features. This problem showed a large improvement in the error
rates: from approximately 40\% for the lasso to less than 20\% for the group lasso.

We provide a generic implementation of the sparse group lasso algorithm in the form of a
\verb!C++! template library. The implementation for multinomial and logistic sparse group lasso
regression is available as an R package. For our implementation the time to compute the
sparse group lasso solution is of the same order of magnitude as the time required for the
multinomial lasso algorithm as implemented in the R-package glmnet. The computation time
of our implementation scales well with the problem size.

\subsection{Sparse group lasso}\label{sec:sgl}
Consider a convex, bounded below and twice continuously differentiable function $f:\mathbb{R}^{n} \to \mathbb{R}$.
We say that $\hat\beta \in \mathbb{R}^n$ is a \emph{sparse group lasso minimizer} if it is a solution to the unconstrained convex optimization problem
\begin{equation}\label{eq:main}
\minimize f + \lambda \Phi
\end{equation}
where $\Phi:\mathbb{R}^{n} \to \mathbb{R}$ is the \emph{sparse group lasso penalty} (defined below) and $\lambda>0$. 

Before defining the sparse group lasso penalty some notation is needed.
We decompose the search space 
\[
\mathbb{R}^{n} = \mathbb{R}^{n_1}\times\dots\times\mathbb{R}^{n_m}
\]
into $m\in \mathbb{N}$ blocks having dimensions $n_i \in \mathbb{N}$ for $i=1,\dots, m$, hence $n = n_1 + \dots + n_m$.
For a vector $\beta \in \mathbb{R}^n$ we write $\beta = (\block{\beta}{1},\dots,\block{\beta}{m}) $ where $\block{\beta}{1} \in \mathbb{R}^{n_1},\dots, \block{\beta}{m} \in \mathbb{R}^{n_m}$. 
For $J = 1,\dots, m$ we call $\block{\beta}{J}$ the $J$'th \emph{block} of $\beta$.
We use the notation $\block{\beta}{J}_i$ to denote the $i$'th coordinate of the $J$'th block of $\beta$, whereas $\beta_i$ is the $i$'th coordinate of $\beta$.

\begin{thm_def}[Sparse group lasso penalty]
The sparse group lasso penalty is defined as
\[
\Phi(\beta) \deff
(1-\alpha)\sum_{J=1}^m\gamma_J\norm{\block{\beta}{J}}_2 +
\alpha\sum_{i=1}^{n}\xi_i\abs{\beta_i}
\]
for $\alpha \in [0,1]$, group weights $\gamma \in [0,\infty)^m$, and
parameter weights $\xi = (\block{\xi}{1},\dots, \block{\xi}{m}) \in
[0,\infty)^n$ where $\block{\xi}{1}\in [0,\infty)^{n_1},\dots,
\block{\xi}{m} \in [0,\infty)^{n_m}$.
\end{thm_def}

The sparse group lasso penalty includes the lasso penalty ($\alpha =
1$) and the group lasso penalty ($\alpha = 0$).  Note also that for
sufficiently large values of $\lambda$ the minimizer of (\ref{eq:main})
is zero. The infimum of these, denoted $\lambda_{\text{max}}$, is computable,
see Section \ref{sec:lambdamax}.

We emphasize that the sparse group lasso penalty is specified by
\begin{itemize}
\item a grouping of the parameters $\beta = (\block{\beta}{1},\dots,\block{\beta}{m})$,
\item and the weights $\alpha, \gamma$ and $\xi$.
\end{itemize} 

In Part \ref{sec:multi} we consider multinomial regression; here the parameter grouping is given by the
multinomial model. For multinomial as well as other regression models, the grouping can also
reflect a grouping of the features.

\part{The multinomial sparse group lasso classifier} \label{sec:multi}

In this section we examine the characteristics of the multinomial sparse group lasso method.
Our main interest is the application of the multinomial sparse group lasso classifier to problems with many classes.
For this purpose we have chosen three classification problems
based on three different data sets, with 10, 18 and 50 classes.  In
\cite{lu2005microrna} the microRNA expression profile of different
types of primary cancer samples is studied.  In Section \ref{sec:mir}
we consider the problem of classifying the primary site based on the
microRNA profiles in this data set.  The Amazon reviews author
classification problem, presented in \cite{liuapplication}, is studied
in Section \ref{sec:ama}.  The messenger RNA profile of different
human muscle diseases is studied in \cite{bakay2006nuclear}.  We
consider, in Section \ref{sec:mus}, the problem of classifying the
disease based on the messenger RNA profiles in this data set.  Table
\ref{tab:datasets} summarizes the dimensions and characteristics of
the data sets and the associated classification problems.  Finally, in
Section \ref{sec:sim}, we examine the characteristics of the method
applied to simulated data sets.

\section{Setup} \label{sec:multinomialsetup}

Consider a classification problem with $K$ classes, $N$ samples, and $p$ features.
Assume given a data set $(x_1, y_1),\dots, (x_N, y_N)$ where, for all $i=1,\dots, N$, $x_i\in\mathbb{R}^p$ is the observed feature vector and $y_i\in \{1,\dots,K\}$ is the categorical response.
We organize the feature vectors in the $N\times p$ \emph{design matrix} 
\[
 X \deff (x_1 \cdots x_N)^T.
\]

As in \cite{friedman2010regularization} we use a symmetric parametrization of the multinomial model.
With $h:\{1,\dots, K\}\times \mathbb{R}^p\to\mathbb{R}$ given by 
\[
h(l, \eta) \deff \frac{\exp(\eta_l)}{\sum_{k=1}^{K}\exp(\eta_k)},
\]
the multinomial model is specified by
\[
P(y_i = l | x_i) = h(l, \block{\beta}{0}+\beta x_i).
\]
The model parameters are organized in the $K$-dimensional vector, $\block{\beta}{0}$, of intercept parameters together with the $K\times p$ matrix
\begin{equation}\label{eq:beta}
\beta \deff \left(\block{\beta}{1} \cdots \block{\beta}{p} \right),
\end{equation}
where $\block{\beta}{i} \in \mathbb{R}^K$ are the parameters associated with the $i$'th feature. 
The lack of identifiability in this parametrization is in practice dealt with by the penalization. 

The log-likelihood is
\begin{equation} \label{eq:multilogl}
\ell(\block{\beta}{0}, \beta) = \sum_{i=1}^N\log
h(y_i, \block{\beta}{0}+\beta x_i).
\end{equation}
Our interest is the sparse group lasso penalized maximum likelihood estimator.
That is, $(\block{\beta}{0},\beta)$ is estimated as a minimizer of the sparse group lasso penalized negative-log-likelihood:  
\begin{equation} \label{eq:multi}
- \ell(\block{\beta}{0},
\beta) + \lambda\left((1-\alpha)\sum_{J=1}^p\gamma_J\norm{\block{\beta}{J}}_2 +
\alpha\sum_{i=1}^{Kp}\xi_i\abs{\beta_i}\right).
\end{equation}
In our applications we let $\gamma_J = \sqrt{K}$ for all $J=1,\dots,
p$ and $\xi_i= 1$ for all $i = 1, \dots, Kp$, but other choices are
possible in the implementation.  Note that the parameter grouping, as part of the penalty specification, is given in terms of the columns in (\ref{eq:beta}), i.e. $m=p$.

\section{Data examples} \label{sec:classsetup}

\begin{table}
 \begin{center}
\begin{tabular}{ l  l  r  r  r }
 Data set & Features & $K$ & $N \ $ & $p \ $ \\
\hline
Cancer sites & microRNA expressions & 18 &162 & 217 \\
Amazon reviews & Various textual features &50 & 1500 & 10k \\
Muscle diseases & Gene expression &10 & 107 &  22k \\
\end{tabular} 
\end{center}
\caption{Summary of data sets and the associated classification problem.}
\label{tab:datasets}
\end{table}

The data sets were preprocessed before applying the multinomial sparse group lasso
estimator. Two preprocessing schemes were used: \emph{normalization} and \emph{standardization}.
Normalization entails centering and scaling of the samples in order to obtain a design matrix
with row means of 0 and row variance of 1. Standardization involves centering and scaling
the features to obtain a design matrix with column means of 0 and column variance of 1.
Note that the order in which normalization and standardization are applied matters.

The purpose of normalization is to remove technical (non-biological) variation. A range of
different normalization procedures exist for biological data. Sample centering and scaling is
one of the simpler procedures. We use this simple normalization procedure for the two
biological data sets in this paper. Normalization is done before and independent from the
sparse group lasso algorithm.

The purpose of standardization is to create a common scale for features. This ensures that
differences in scale will not influence the penalty and thus the variable selection.
Standardization is an option for the sparse group lasso implementation, and it is applied as
the last preprocessing step for all three example data sets.

We want to compare the performance of the multinomial sparse
group lasso estimator for different values of the regularization
parameter $\alpha$.  Applying the multinomial sparse group lasso
estimator with a given $\alpha \in [0,1]$ and $\lambda$-sequence,
$\lambda_1, \ldots, \lambda_d > 0$, results in a sequence of estimated
models with parameters $\{\hat\beta(\lambda_i, \alpha)\}_{i=1,\dots,d}$. 
The generalization error can be estimated by cross validation \cite{HastTibsFrie2001}.
For our applications we keep the sample ratio between classes in the cross validation subsets approximately fixed to that of the entire data set.
Hence, we may compute a sequence, $\{\widehat\Err(\lambda_i, \alpha)\}_{i=1, \dots, d}$, of estimated expected generalization errors for the sequence of 
models. However, for given $\alpha_1$ and $\alpha_2$ we cannot simply compare $\widehat\Err(\lambda_i,
\alpha_1)$ and $\widehat\Err(\lambda_i, \alpha_2)$, since the $\lambda_i$
value is scaled differently for different values of $\alpha$.  We
will instead compare the models with the same number of non-zero
parameters and the same number of non-zero parameter groups,
respectively. Define
\[
 \hat\Theta(\lambda, \alpha) \deff \sum_{J=1}^p I(\block{\hat\beta}{J}(\lambda,\alpha) \neq 0)
\]
with $\hat\beta(\lambda, \alpha)$ the estimator of $\beta$ for the given values of $\lambda$ and $\alpha$.
That is, $\hat\Theta(\lambda, \alpha)$ is the number of non-zero parameter blocks in the fitted model. 
Note that there is a one-to-one correspondence between parameter blocks and features in the design matrix.
Furthermore, we define the total number of non-zero parameters as
\[
 \hat\Pi(\lambda, \alpha) \deff \sum_{i=1}^n I(\hat\beta_i(\lambda, \alpha) \neq 0).
\]

In particular, we want to compare the fitted models with the
same number of parameter blocks.  There may, however, be more than one
$\lambda$-value corresponding to a given value of $\hat\Theta$. 
Thus we compare the models on a subsequence of the $\lambda$-sequence.
With $\theta_1 < \dots < \theta_{d'}$ for $d' \le d$ denoting the different elements of the set $\{\hat\Theta(\lambda_i, \alpha)\}_{i =1, \ldots, d}$ in increasing order we define 
\[
\tilde\lambda_i(\alpha) \deff \min\set{\lambda}{\hat\Theta(\lambda, \alpha) = \theta_i}.
\]
We then compare the characteristics of the multinomial sparse group
lasso estimators for different $\alpha$ values by comparing the estimates 
\[
\left\{\left(\widehat\Err(\tilde\lambda_i(\alpha), \alpha), \hat\Theta(\tilde\lambda_i(\alpha)), \hat\Pi(\tilde\lambda_i(\alpha))\right)\right\}_{i=1,\dots, d'}.
\]

\subsection{Cancer sites} \label{sec:mir}

\begin{figure}
\centering
\includegraphics{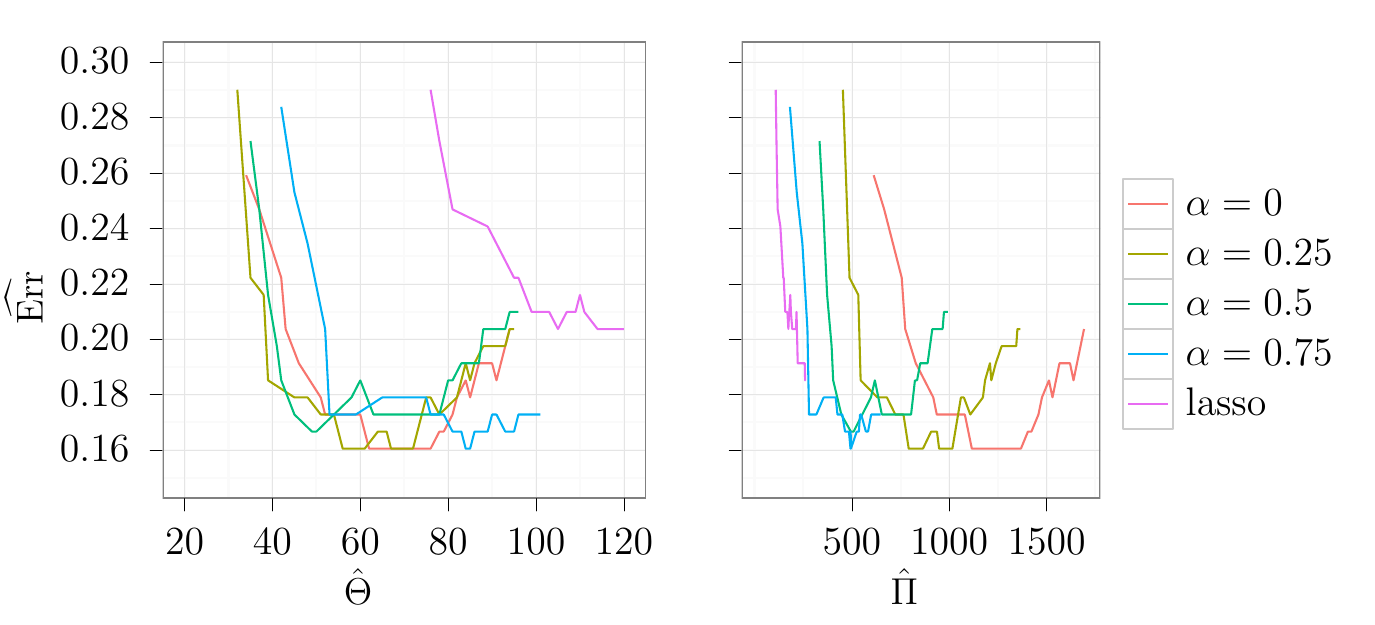}
\caption{Estimated expected generalization error, for different values
  of $\alpha$, for the microRNA cancer sites data set.  The cross
  validation based estimate of the expected misclassification error is
  plotted against the number of non-zero parameter blocks in the model
  (left), and against the number of non-zero parameters in the model
  (right). The estimated standard error is approximately 0.03 for all models.} \label{fig:mir}
\end{figure}

The data set consists of bead-based expression data for 217 microRNAs
from normal and cancer tissue samples.  The samples are divided into
11 normal classes, 16 tumor classes and 8 tumor cell line classes.
For the purpose of this study we select the normal and tumor classes
with more than 5 samples.  This results in an 18 class data set with
162 samples.  The data set is unbalanced, with the number of samples in
each class ranging from 5 to 26 and with an average of 9 samples per
class.  Data was normalized and then standardized before running the
sparse group lasso algorithm.  For more information about this data
set see \cite{lu2005microrna}.  The data set is
available from the Gene Expression Omnibus with accession number
GSE2564.

Figure \ref{fig:mir} shows the result of a 10-fold cross validation
for 5 different values of $\alpha$, including the lasso and group
lasso.  The $\lambda$-sequence runs from $\lambda_{\text{max}}$ to
$10^{-4}$, with $d=200$.  It is evident that the group lasso and
sparse group lasso models achieve a lower expected error using fewer
genes than the lasso model.  However, models with a low $\alpha$ value
have a larger number of non-zero parameters than models with a high
$\alpha$ value.  A reasonable compromise could be the model with
$\alpha=0.25$.  This model does not only have a low estimated expected
error, but the low error is also achieved with a lower estimated number of
non-zero parameters, compared to group lasso.

\subsection{Amazon reviews} \label{sec:ama}

\begin{figure}
\centering
\includegraphics{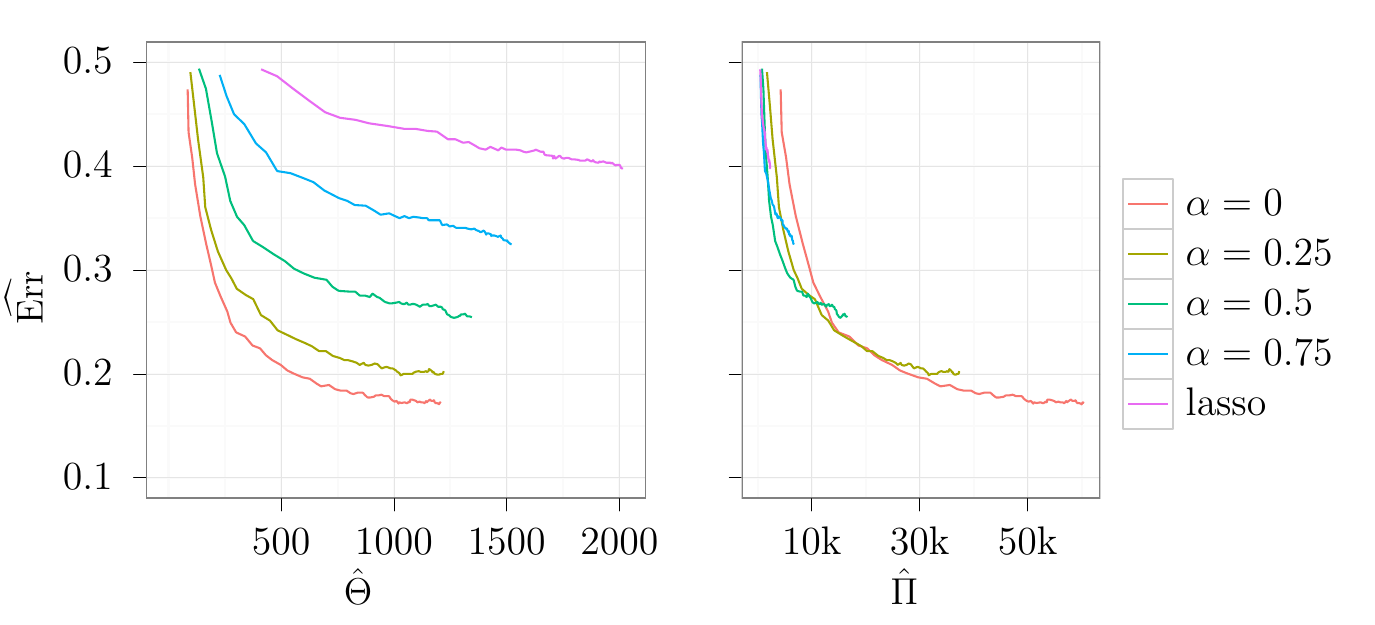}
  \caption{Estimated expected generalization error, for different values of $\alpha$, for the Amazon reviews author classification problem.
The cross validation based estimate of expected misclassification error is plotted against the number of non-zero parameter blocks in the model (left), and against the number of non-zero parameters in the model (right).
The estimated standard error is approximately 0.01 for all models.}\label{fig:amazone}
\end{figure}

The Amazon review data set consists of 10k textual features (including
lexical, syntactic, idiosyncratic and content features) extracted from
1500 customer reviews from the Amazon Commerce Website.  The reviews were
collected among the reviews from 50 authors with 50 reviews per author.  The primary
classification task is to identify the author based on the textual
features.  The data and feature set were presented in
\cite{liuapplication} and can be found in the UCI machine learning
repository \cite{Frank+Asuncion:2010}.  In \cite{liuapplication} a
Synergetic Neural Network is used for author classification, and a
$2$k feature based 10-fold CV accuracy of 0.805 is reported.  The
feature selection and training of the classifier were done separately.

We did 10-fold cross validation using multinomial sparse group lasso
for five different values of $\alpha$.  The results are shown in
Figure \ref{fig:amazone}.  The $\lambda$-sequence runs from
$\lambda_{\text{max}}$ to $10^{-4}$, with $d=100$.  
The design matrix is sparse for this data set. Our implementation of the multinomial
sparse group lasso algorithm utilizes the sparse design matrix to gain speed and for memory
efficiency. No normalization was applied for this data set. Features were scaled to have
variance 1, but were not centered.

For this data set it is evident that lasso performs badly and that the group lasso performs best - in fact much better than lasso. 
The group lasso achieves an accuracy of around 0.82 with a feature set of size $\sim 1$k.
This outperforms the neural network in \cite{liuapplication}.

\subsection{Muscle diseases}\label{sec:mus}

\begin{figure}
\centering
\includegraphics{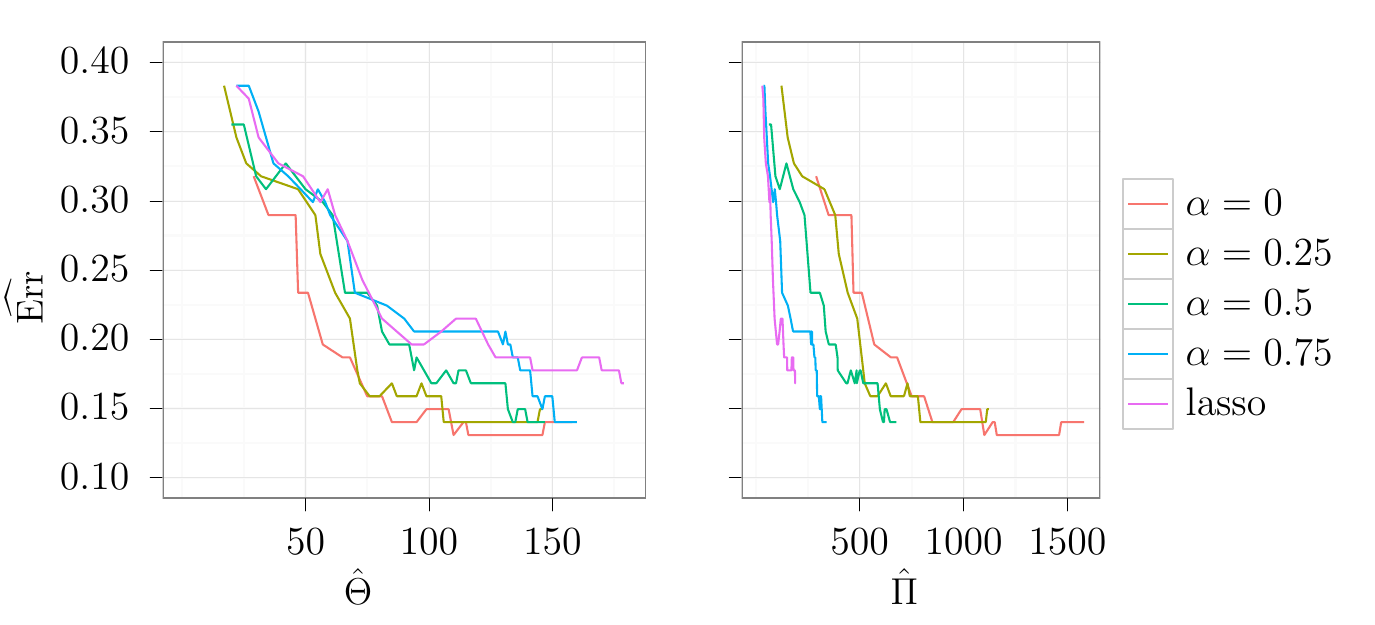}
\caption{Estimated expected generalization error, for different values
  of $\alpha$, for the muscle disease classification problem.  The
  cross validation based estimate of expected misclassification error
  is plotted against the number of non-zero parameter blocks in the
  model (left), and against the number of non-zero parameters in the
  model (right) The estimated standard error is approximately 0.04 for all models.}\label{fig:mus}
\end{figure}

This data set consists of messenger RNA array expression data of 119
muscle biopsies from patients with various muscle diseases.  The
samples are divided into 13 diagnostic groups.  For this study we only
consider classes with more than 5 samples.  This results in a
classification problem with 107 samples and 10 classes.  The data set
is unbalanced with class sizes ranging from 4 to 20 samples per
class.  Data was normalized and then standardized before running the
sparse group lasso algorithm.  For background information on this data
set, see \cite{bakay2006nuclear}. The data set is available from the
Gene Expression Omnibus with accession number GDS1956.

The results of a 10-fold cross validation are shown in Figure \ref{fig:mus}.
The $\lambda$-sequence runs from $\lambda_{\text{max}}$ to $10^{-5}$, with $d=200$.
We see the same trend as in the other two data examples.
Again the group lasso models perform well, however not
significantly better than the closest sparse group lasso models ($\alpha = 0.25$).
The lasso models perform reasonably well on this data set, but they are
still outperformed by the sparse group lasso models.

\section{A simulation study}\label{sec:sim}

In this section we investigate the characteristics of the sparse group
lasso estimator on simulated data sets.  We are primarily interested
in trends in the generalization error as $\alpha$ is varied and
$\hat\lambda$ is selected by cross validation on a relatively small
training set.  We suspect that this trend will depend on the
distribution of the data.  We restrict our attention to multiclass
data where the distribution of the features given the class is
Gaussian.  Loosely speaking, we suspect that if the differences in the
data distributions are very sparse, i.e. the centers of the
Gaussian distributions are mostly identical across classes, the lasso will produce models with the lowest generalization error.
If the data distribution is sparse, but not very sparse, then the optimal $\alpha$
is in the interval $(0,1)$. 
For a dense distribution, typically with differences being among all classes, we expect the
group lasso to perform best. The simulation study confirms this.

The mathematical formulation is as follows. Let
\[
 \mu = (\mu_1 \dots \mu_K)
\]
where $\mu_i \in \mathbb{R}^p$ for $i=1,\dots, K$ and $p=p_a+p_b$.
Denote by $\mathcal{D}_\mu$ a data set consisting of $N$ samples for
each of the $K$ classes -- each sampled from the Gaussian distribution
with centers $\mu_1, \dots, \mu_K$, respectively, and with a common
covariance matrix $\Sigma$.  Let $\hat\lambda$ be the smallest 
$\lambda$-value with the minimal estimated expected generalization
error, as determined by cross validation on $\mathcal{D}_\mu$.  Denote
by $\Err_\mu(\lambda, \alpha)$ the generalization error of the model
$\hat\beta(\lambda, \alpha)$ that has been estimated from the training
set $\mathcal{D}_\mu$, by the sparse group lasso, for the given values
of $\lambda$ and $\alpha$.  Then let
\[
 \textstyle Z_\mu(\alpha) = \Err_\mu(\hat\lambda, \alpha)- \Err_{\text{Bayes}}(\mu)
\]
where $\Err_{\text{Bayes}}(\mu)$ is the Bayes rate.  We are interested
in trends in $Z_\mu$, as a function of $\alpha$, for different
configurations of $\mu_1, \dots, \mu_K$.  To be specific, we will
sample $\mu_1, \dots, \mu_K$ from one of the following distributions:
\begin{itemize}
\item A \emph{sparse} model distribution, where the first $p_a$
  entries of $\mu_i$ are i.i.d. with a distribution that is a mixture
  of the uniform distribution on $[-2,2]$ and the degenerate
  distribution at 0 with point probability $p_0$.
\item A \emph{dense} model distribution, where the first $p_a$ entries of $\mu_i$ are i.i.d. Laplace distributed with location 0 and scale $b$.
\end{itemize}
The last $p_b$ entries are zero.  We take $p_a =
\lfloor5/(1-p_0)\rfloor$ throughout for the sparse model distribution.
The within class covariance matrix $\Sigma$ is constructed using
features from the cancer site data set.  Let $\Sigma_0$ be the
empirical covariance matrix of $p$ randomly chosen features.  To avoid that the
covariance matrix become singular we take
\[
 \Sigma = (1-\delta)\Sigma_0 + \delta\mathbf{I}
\]
for $\delta \in (0,1)$.

The primary quantity of interest is
\begin{equation}\label{eq:simerr}
\err(\alpha) \deff \E \left(Z_\mu(\alpha)\right), 
\end{equation}
the expectation being over $\mu$ and the data set $\mathcal{D}_\mu$.
We are also interested in how well we can estimate the non-zero patterns of the $\mu_i$'s.
Consider this as $Kp$ two class classification problems, one for each parameter, where we predict the
$\mu_{ij}$ to be non-zero if $\hat \beta_{ij}$ is non-zero, and
$\mu_{ij}$ to be zero otherwise.  
We calculate the number of false positives, true positives, false negatives and true negatives.
The positive predictive value (ppv) and the true positive rate (tpr) are of particular interest.  The true positive
rate measures how sensitive a given method is at discovering non-zero
entries.  The positive predictive value measures the precision with
which the method is selecting the non-zero entries. 
We consider the following two quantities
\begin{equation} \label{eq:simtpr}
\Tpr(\alpha) \deff \E\left[\Tpr\left(\hat\beta(\hat\lambda, \alpha)\right)\right] \text{ and } \Ppv(\alpha) \deff \E\left[\Ppv\left(\hat\beta(\hat\lambda, \alpha)\right)\right]. 
\end{equation}
\begin{figure}
  \centering
\includegraphics{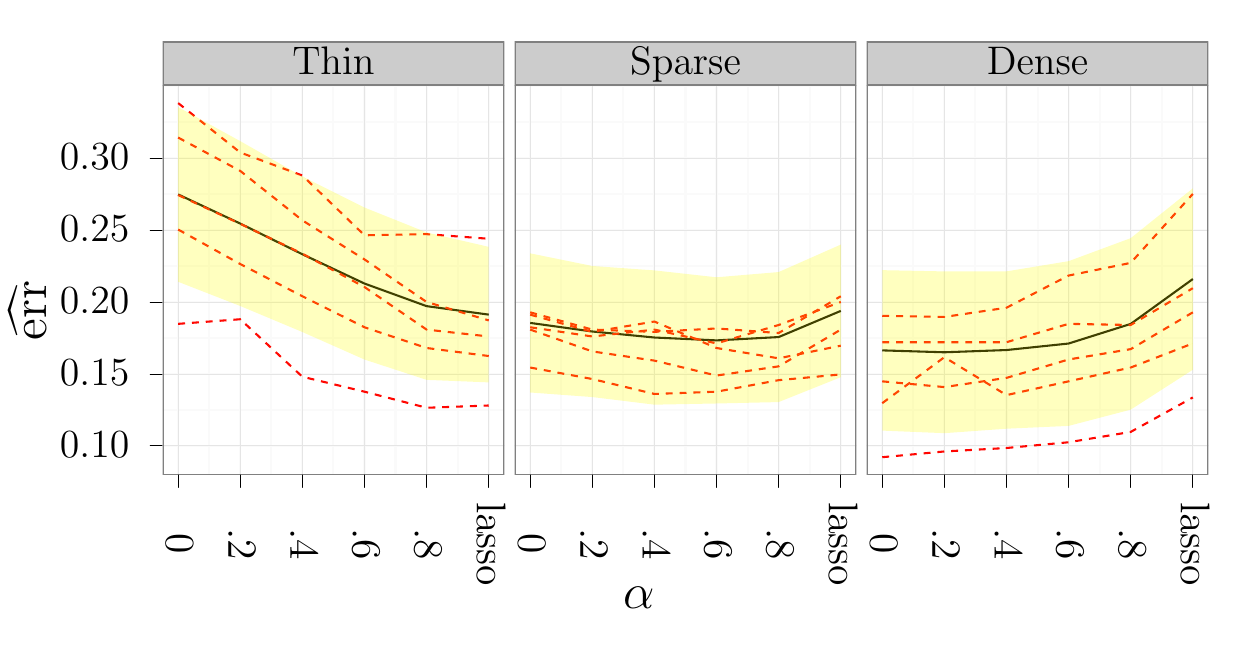}
  \caption{The estimated expected error gab (solid black line) for the three configurations. 
The central 95\% of the distribution of $Z_{\mu}(\alpha)$ is shown as the shaded area on the plot.
The error gab for 5 randomly selected $\mu$-configurations is shown (red dashed lines).}
  \label{fig:simerr}
\end{figure}

In order to estimate the quantities (\ref{eq:simerr}) and
(\ref{eq:simtpr}) we sample $M$ configurations of  $\mu$ from one of the
above distributions.  For each configuration we sample a training and
a test data set of sizes $NK$ and $100K$, respectively.  Using the
training data set we fit the model $\hat\beta(\hat\lambda,\alpha)$
and estimate $Z_\mu(\alpha)$ using the test data set. Estimates
$\widehat\err(\alpha)$, $\widehat\Tpr(\alpha)$ and
$\widehat\Ppv(\alpha)$ are the corresponding averages over the
$M$ configurations.

For this study we chose $M=100$, $N=15$, $K=25$, $p_b=50$, $\delta = 0.25$ and the following three configuration distributions:
\begin{itemize}
\item \emph{Thin} configurations, where the centers are distributed according to the sparse model distribution with $p_0=0.95$, as defined above.
\item \emph{Sparse} configurations, where the centers are distributed according to the sparse model distribution with $p_0=0.80$.
\item \emph{Dense} configurations,  where the centers are distributed according to the dense model distribution with scale $b=0.2$ and $p_a=25$. 
\end{itemize}

In Figure \ref{fig:simerr} we see that for thin configurations the lasso has the lowest estimated error gab, along with the sparse group lasso with $\alpha=0.8$.
For the sparse configurations the results
indicate that the optimal choice of $\alpha$ is in the open interval $(0,1)$, but in this
case all choices of $\alpha$ result in a comparable error gab. For
the dense configurations the group lasso is among the methods with the
lowest error gab.

\begin{figure}
  \centering
\includegraphics{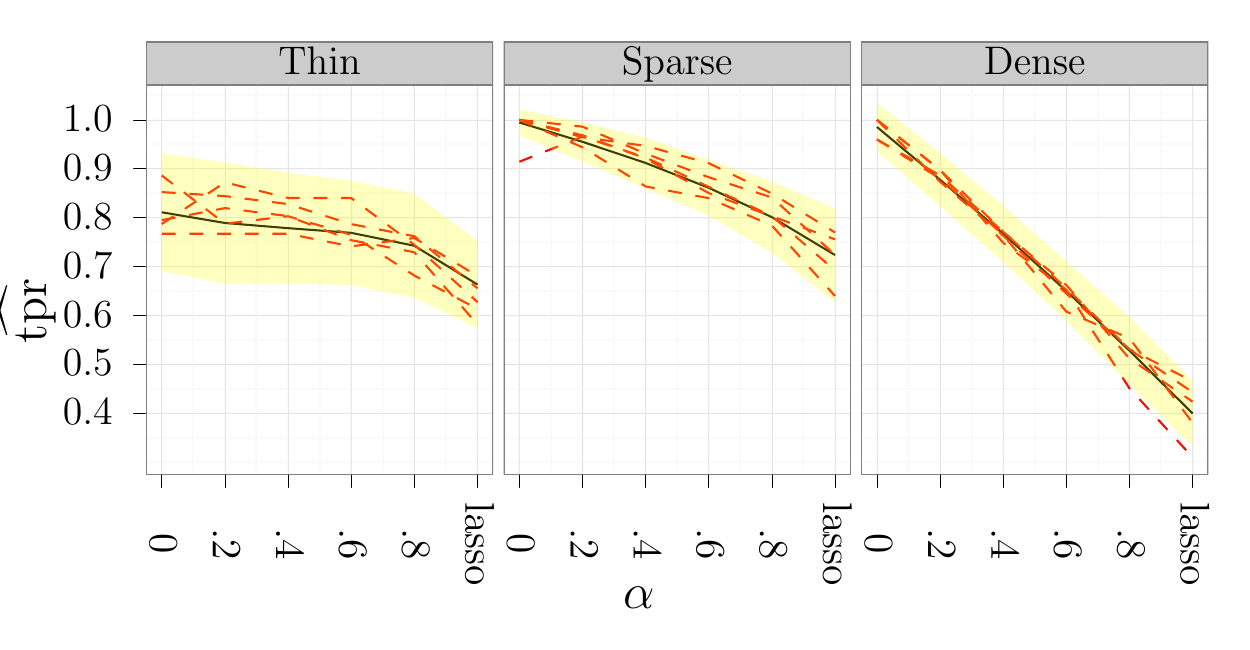}
  \caption{The estimated expected true positive rate (solid black line) for the three configurations. 
The central 95\% of the distribution of $\Tpr$ is shown as the shaded area on the plot.
The true positive rate for 5 randomly selected $\mu$-configurations is shown (red dashed lines).} \label{fig:simtpr}
\end{figure}

\begin{figure}
  \centering
\includegraphics{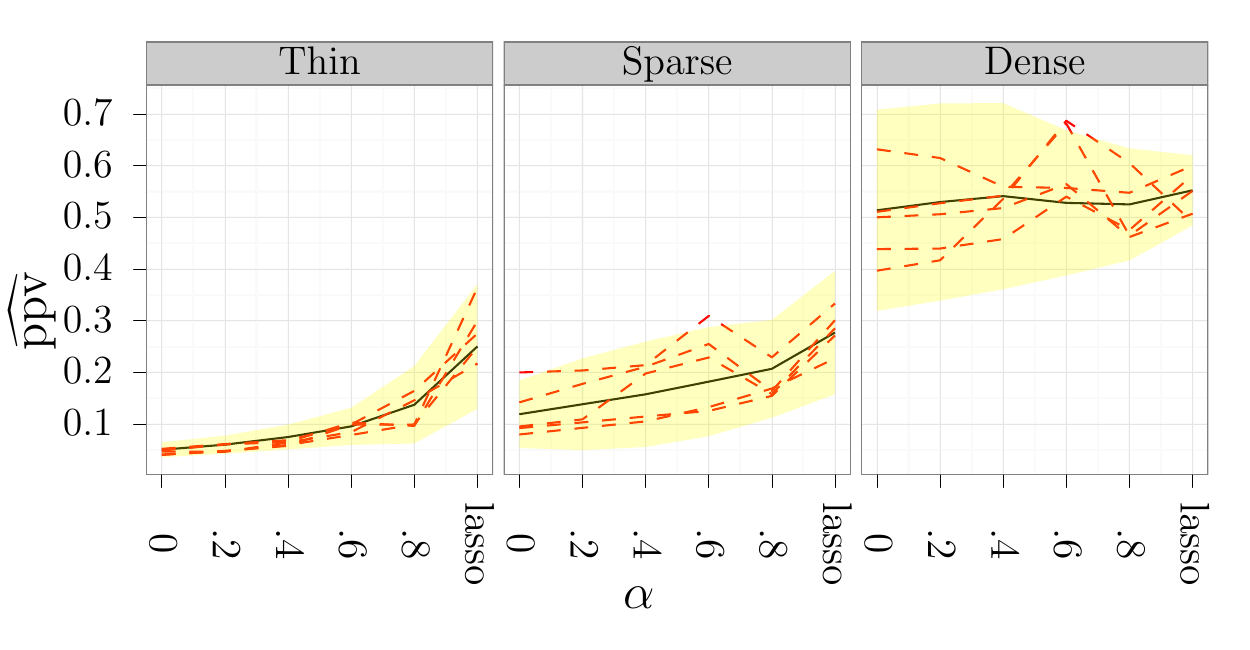}
  \caption{The estimated expected positive predictive value (solid black line) for the three configurations. 
The central 95\% of the distribution of $\Ppv$ is shown as the shaded area on the plot.
The positive predictive value for 5 randomly selected $\mu$-configurations is shown (red dashed lines).} \label{fig:simppv}
\end{figure}

In Figure \ref{fig:simtpr} we plotted the true positive rate for
the three configurations.  Except for the thin configurations, the
lasso is markedly less sensitive than the sparse group and group
lasso methods.  However, looking at Figure \ref{fig:simppv} we see
that the sparse group and group lasso methods have a lower 
precision than the lasso, except for the dense configurations.
We note that the group lasso has the worst precision,
except for the dense configurations.

\part{The sparse group lasso algorithm} \label{sec:algo}

In this section we present the sparse group lasso algorithm.  The
algorithm is applicable to a broad class of loss functions. Specifically, we require that the loss function $f:\mathbb{R}^{n} \to
\mathbb{R}$ is convex, twice continuously differentiable and bounded
below.  Additionally, we require that all quadratic approximations
around a point in the sublevel set
\[
\set{\beta \in \mathbb{R}^n}{f(\beta)+\lambda\Phi(\beta) \le f(\beta_0)+\lambda\Phi(\beta_0) }
\]
are bounded below, where $\beta_0 \in \mathbb{R}^n$ is the initial point.
The last requirement will ensure that all subproblems are well defined. 

The algorithm solves (\ref{eq:main}) for a decreasing sequence of
$\lambda$ values ranging from $\lambda_{\text{max}}$ to a user specified
$\lambda_{\text{min}}$.  The algorithm consists of four nested main
loops:
\begin{itemize}
\item A numerical continuation loop, decreasing $\lambda$.
\item An outer coordinate gradient descent loop (Algorithm \ref{algo:outer}).
\item A middle block coordinate descent loop (Algorithm \ref{algo:middle}).
\item An inner modified coordinate descent loop (Algorithm \ref{algo:inner}). 
\end{itemize}

In Section \ref{sec:outer} to \ref{sec:inner} we discuss the outer,
middle and inner loop, respectively.  In Section \ref{sec:ubound} we
develop a method allowing us to bypass computations of large parts of
the Hessian, hereby improving the performance of the middle loop.
Section  \ref{sec:soft} provides a discussion of the available software solutions, as well as run-time performance of the current implementation.

The theoretical basis of the optimization methods we apply can be found in \cite{tseng2001convergence, tseng2009coordinate}.
A short review tailored to this paper is given in \ref{app:bcd}.

\subsection{The sparse group lasso penalty}
In this section we derive fundamental results regarding the sparse group lasso penalty.

We first observe that $\Phi$ is separable in the sense that if, for any group
$J\in 1,\dots, m$, we define the
convex penalty $\block{\Phi}{J}:\mathbb{R}^{n_J} \to \mathbb{R}$ by
\[
\block{\Phi}{J}(\hat x) \deff (1-\alpha)\gamma_J\norm{\hat x}_2 + \alpha\sum_{i=1}^{n_J}\block{\xi}{J}_i\abs{\hat x_i}
\]
then $\Phi(\beta) = \sum_{J = 1}^m \block{\Phi}{J}(\block{\beta}{J})$.
Separability of the penalty is required to ensure convergence of coordinate descent methods, see \cite{tseng2009coordinate,
  tseng2001convergence}, and see also \ref{app:bcd}.

In a block coordinate descent scheme the primary minimization problem is solved by minimizing each block, one at a time, until convergence.
We consider conditions ensuring that 
\begin{equation}\label{eq:subg}
0 \in \argmin{x\in\mathbb{R}^{n_J}} g(x) + \lambda \block{\Phi}{J}(x)
\end{equation}
for a given convex and twice continuously differentiable function $g:\mathbb{R}^{n_J}\to \mathbb{R}$.
For $J=1, \dots, m$ a straightforward calculation shows that the subgradient
of $\block{\Phi}{J}$ at zero is
\[
\partial\block{\Phi}{J}(0) = (1-\alpha)\gamma_JB^{n_J}+\alpha\diag(\block{\xi}{J})T^{n_J}
\]
where $B^n \deff \set{x\in \mathbb{R}^n}{\norm{x}_2\le 1}$, $T^n \deff [-1,1]^n$ and where for $x\in \mathbb{R}^n$ $\diag(x)$ denotes the $n\times n$ diagonal matrix with diagonal $x$.
For an introduction to the theory of subgradients see Chapter 4 in \cite{bertsekas2003convex}.

Proposition \ref{thm:kprop} below gives a necessary and sufficient condition for (\ref{eq:subg}) to hold.
Before we state the proposition the following definition is needed.

\begin{thm_def}
For $n\in \mathbb{N}$ we define the map $\kappa : \mathbb{R}^n\times
\mathbb{R}^n \to \mathbb{R}^n$ by 
\[
 \kappa(v, z)_i \deff \begin{cases}
0 & \abs{z_i} \le v_i \\
z_i - \sgn(z_i)v_i & \text{otherwise}
\end{cases} \text{ for $i = 1, \dots, n$}
\]
and the function $K : \mathbb{R}^n\times \mathbb{R}^n \to \mathbb{R}$ by
\[
K(v, z) \deff \norm{\kappa(v,z)}_2^2 = \sum_{\set{i}{\abs{z_i} >
v_i}} \left(z_i - \sgn(z_i)v_i\right)^2.
\]
\end{thm_def}

\begin{thm_prop} \label{thm:kprop}
Assume given $a > 0$, $v, z\in \mathbb{R}^n$ and define the closed sets
\[
Y = z + \diag(v)T_n \quad \text{and} \quad  X = aB^n + Y.
\]
Then the following hold: 
\begin{itemize}
\item[\itemdef{a}] $\kappa(v, z) = \argmin{y \in Y} \norm{y}_2$.
\item[\itemdef{b}] $0 \in X$ if and only if $K(v, z) \le a^2$.
\item[\itemdef{c}] If $K(v, z) > a^2$ then $\argmin{x \in X} \norm{x}_2 = 
\left(1-a/\sqrt{K(v,z)}\right)\kappa(v,z)$.
\end{itemize}
\end{thm_prop}
The proof of Proposition \ref{thm:kprop} is given in \ref{sec:proofkprop}.
Proposition \ref{thm:kprop} implies that (\ref{eq:subg}) holds if and only if 
\[
\sqrt{K(\lambda\alpha\block{\xi}{J}, \nabla g(0))} \leq \lambda (1-\alpha)
\gamma_J.
\]

The following observations will prove to be valuable.
Note that we use $\preceq$ to denote coordinatewise ordering.
\begin{thm_lemma}\label{thm:klemma}
For any three vectors $v,z, z' \in \mathbb{R}^{n}$ the following hold:
\begin{itemize}
\item[\itemdef{a}] $K(v, z) = K(v, \abs{z})$.
\item[\itemdef{b}] $K(v, z) \le K(v, z')$ when $\abs{z} \preceq \abs{z'}$.
\end{itemize}

\end{thm_lemma}
\begin{proof}
\itemref{a} is a simple calculation and \itemref{b} is a consequence of the definition
and \itemref{a}.
\end{proof}

\subsection{The $\lambda$-sequence} \label{sec:lambdamax}

For sufficiently large $\lambda$ values the only solution to (\ref{eq:main}) will be zero.
We denote the infimum of these by $\lambda_{\text{max}}$.
By using the above observations it is clear that
\begin{align*}
 \lambda_{\text{max}} &\deff \inf\set{\lambda > 0}{\hat\beta(\lambda) = 0} \\
&=\inf\set{\lambda > 0}{\forall J=1,\dots,m : \sqrt{K(\lambda\alpha\block{\xi}{J},
\block{\nabla f(0)}{J})} \le
\lambda(1-\alpha)\gamma_J} \\
&=\max_{J=1,\dots,m} \inf \set{\lambda > 0}{\sqrt{K(\lambda\alpha\block{\xi}{J},
\block{\nabla f(0)}{J})}\le
\lambda(1-\alpha)\gamma_J}.
\end{align*}

It is possible to compute
\[
 \inf \set{\lambda > 0}{\sqrt{K(\lambda\alpha\block{\xi}{J}, \block{\nabla f(0)}{J})}\le \lambda(1-\alpha)\gamma_J}
\]
by using the fact that the function $\lambda \to K(\lambda\alpha\block{\xi}{J}, \block{\nabla f(0)}{J})$ is piecewise quadratic and monotone.

\subsection{Outer loop} \label{sec:outer}
In the outer loop a coordinate gradient descent scheme is used. In this paper we use the
simplest form of this scheme. In this simple form the coordinate gradient descent method is
similar to Newton's method; however the important difference is the way the non-differentiable penalty is handled. 
The convergence of the coordinate gradient descent method is not trivial and is established in \cite{tseng2009coordinate}.

The algorithm is based on a quadratic approximation of the loss function $f$, at the current estimate of the minimizer.
The difference, $\Delta$, between the minimizer of the penalized quadratic
approximation and the current estimate is then a descent direction.  A
new estimate of the minimizer of the objective is found by
applying a line search in the direction of $\Delta$.  We repeat this
until a stopping condition is met, see Algorithm \ref{algo:outer}.
Note that a line search is necessary in order to ensure global
convergence.  For most iterations, however, a step size of 1 will give sufficient
decrease in the objective.  With $q = \nabla f(\beta)$ and $H =
\nabla^2f(\beta)$ the quadratic approximation of $f$ around the
current estimate, $\beta$, is
\begin{multline*}
q^T(x-\beta) + \frac{1}{2}(x-\beta)^TH(x-\beta) \\ = q^Tx - q^T\beta + \frac{1}{2}x^THx - \frac{1}{2}\left(\beta^THx+x^TH\beta\right) + \frac{1}{2}\beta^TH\beta.
\end{multline*}
$H$ is symmetric, thus it follows that the quadratic approximation of $f$ around $\beta$ equals
\begin{equation*}
Q(x) -  q^T\beta + \frac{1}{2}\beta^TH\beta,
\end{equation*}
where $Q:\mathbb{R}^n \to\mathbb{R}$ is defined by
\[
Q(x) \deff (q - H\beta)^Tx+\frac{1}{2}x^THx.
\]
We have reduced problem (\ref{eq:main}) to the following penalized quadratic optimization problem
\begin{equation}\label{eq:quadopt}
\underset{x \in \mathbb{R}^n}{\min} \, Q(x) +  \lambda \Phi(x).
\end{equation}

\begin{algorithm} 
\caption{Outer loop. Solve (\ref{eq:main}) by coordinate gradient descent.}
\begin{algorithmic} \label{algo:outer}
\REQUIRE $\beta = \beta_0$
\REPEAT
\STATE Let $q = \nabla f(\beta)$, $H = \nabla^2f(\beta)$ and $Q(x) = (q -
H\beta)^Tx+\frac{1}{2}x^THx$.
\STATE Compute $\hat\beta = \underset{x \in
\mathbb{R}^n}{\arg\min} \, Q(x) +  \lambda
\Phi(x)$.
\STATE Compute step size $t$ and set $\beta = \beta + t\Delta$, for $\Delta=\beta-\hat\beta$. 
\UNTIL{stopping condition is met.}
\end{algorithmic}
\end{algorithm}

The convergence of Algorithm \ref{algo:outer} is implied by Theorem 1e in \cite{tseng2009coordinate}. This implies: 

\begin{thm_prop}
Every cluster point of the sequence $\{\beta_k\}_{k\in\mathbb{N}}$ generated by Algorithm \ref{algo:outer} is a solution of problem (\ref{eq:main}).
\end{thm_prop}

\begin{thm_rem}
The convergence of Algorithm \ref{algo:outer} is ensured even if $H$ is a (symmetric) positive definite matrix approximating $\nabla^2f(\beta)$. 
For high dimensional problems it might be computationally beneficial
to take $H$ to be diagonal, e.g. as the diagonal of $\nabla^2f(\beta)$.
\end{thm_rem}

\subsection{Middle loop}

In the middle loop the penalized quadratic optimization problem (\ref{eq:quadopt}) is solved. 
The penalty $\Phi$ is block separable, i.e.
\[
Q(x) +  \lambda \Phi(x) = Q(x) + \lambda \sum_{J=0}^p\block{\Phi}{J}(\block{x}{J})
\] 
with $\block{\Phi}{J}$ convex, and we can therefore use the block coordinate descent method over the blocks $\block{x}{1},\dots,\block{x}{m}$. 
The block coordinate descent method will converge to a minimizer even for non-differentiable objectives if the non-differentiable parts are block separable, see \cite{tseng2001convergence}. 
Since $\Phi$ is separable and $Q$ is convex, twice continuously differentiable and bounded below, the block coordinate descent scheme converges to the minimizer of problem (\ref{eq:quadopt}). 
Hence, our problem is reduced to the following collection of problems, one for each $J= 1, \dots, m$, 
\begin{equation}\label{eq:middle}
\underset{\hat x \in
\mathbb{R}^{n_J}}{\min} \, \block{Q}{J}(\hat x) +  \lambda \block{\Phi}{J}(\hat x)
\end{equation}
where $\block{Q}{J}: \mathbb{R}^{n_J} \to \mathbb{R}$ is the quadratic function 
\[\hat x \to
Q(\block{x}{1},\dots, \block{x}{J-1}, \hat x, \block{x}{J+1}, \dots,
\block{x}{m})
\] 
up to an additive constant. We decompose an $n \times n$ matrix $H$ into block matrices in the following way
\[
H = \begin{pmatrix}
                  H_{11} & H_{12} & \cdots & H_{1m} \\
                  H_{21} & H_{22} & \cdots & H_{2m}\\
		  \vdots  & \vdots  & \ddots & \vdots  \\
                  H_{m1} & H_{m2} & \cdots & H_{mm}\\
                 \end{pmatrix}
\]
where $H_{IJ}$ is an $n_I \times n_J$ matrix. By the symmetry of $H$
it follows that
\begin{align*}
\block{Q}{J}(\hat x) &= \hat x ^T \block{(q-H\beta)}{J} + \frac{1}{2}\left(2 \sum_{I}\hat x^T H_{JI}\block{x}{I}-\hat x^TH_{JJ}\block{x}{J}+\hat x^TH_{JJ}\hat x\right) \\
&= \hat x^T \left(\block{q}{J} + \block{[H(x-\beta)]}{J} - H_{JJ}\block{x}{J}\right) + \frac{1}{2}\hat x^TH_{JJ}\hat x
\end{align*}
up to an additive constant.
We may, therefore, redefine
\[
\block{Q}{J}(\hat x) \deff \hat x^T\block{g}{J} + \frac{1}{2} \hat x^T H_{JJ}\hat x
\]
where the \emph{block gradient} $\block{g}{J}$ is defined by
\begin{equation}\label{eq:blockgradient}
\block{g}{J} \deff \block{q}{J} + \block{\left[H(x-\beta)\right]}{J} - H_{JJ}\block{x}{J}.
\end{equation}

For the collection of problems given by (\ref{eq:middle}) a considerable fraction of the minimizers will be zero in practice. 
By Lemma \ref{thm:kprop} this is the case if and only
if
\[
\sqrt{K(\lambda\alpha\block{\xi}{J}, \block{g}{J})} \leq \lambda
(1-\alpha) \gamma_J.
\]

These considerations lead us to Algorithm \ref{algo:middle}.

\begin{algorithm}
\caption{Middle loop. Solve (\ref{eq:quadopt}) by block coordinate descent.}
\begin{algorithmic}\label{algo:middle}
\REPEAT
\STATE Choose next block index $J$ according to the cyclic rule.
\STATE Compute the block gradient $\block{g}{J}$.
\IF{ $\sqrt{K(\lambda\alpha\block{\xi}{J}, \block{g}{J})} \leq
\lambda(1-\alpha)\gamma_J$} 
\STATE Let $\block{x}{J} = 0$.
\ELSE
\STATE Let $\block{x}{J} = \underset{\hat x \in
\mathbb{R}^{n_J}}{\arg\min}\,\block{Q}{J}(\hat x) + \lambda \block{\Phi}{J}(\hat
x)$.
\ENDIF
\UNTIL{stopping condition is met.}
\end{algorithmic}
\end{algorithm}

\subsection{Inner loop} \label{sec:inner}

Finally we need to determine the minimizer of (\ref{eq:middle}), i.e. the minimizer of
\begin{equation} \label{eq:inner}
\underbrace{\vphantom{\sum_{i=0}^{n_J}}\block{Q}{J}(\hat x) + \lambda (1-\alpha)\gamma_J\norm{\hat x}_2}_{\text{loss}} + \underbrace{\alpha\sum_{i=0}^{n_J}\block{\xi}{J}_i\abs{\hat x_i}}_{\text{penalty}}.
\end{equation}

The two first terms of (\ref{eq:inner}) are considered the loss function and the last term is the penalty.
Note that the loss is not differentiable at zero (due to the $L_2$-norm), thus we cannot completely separate out the non-differentiable parts.
This implies that ordinary block coordinate descent is not guaranteed to converge to a minimizer. 
Algorithm \ref{algo:inner} adjusts for this problem, and we have the following proposition. 

\begin{thm_prop}
For any $\epsilon > 0$ the cluster points of the sequence $\{\hat x_k\}_{k\in\mathbb{N}}$ generated by Algorithm \ref{algo:inner} are minimizers of (\ref{eq:inner}).
\end{thm_prop}
\begin{proof}
Since $\block{Q}{J}(0) + \lambda \block{\Phi}{J}(0) = 0$ Algorithm \ref{algo:inner} is a modified block coordinate descent scheme. 
Furthermore $J$ is chosen such that (\ref{eq:inner}) is not optimal at 0.
We can therefore apply Lemma \ref{thm:modbcd} in \ref{app:modbcd}, from which the claim follows directly.
\end{proof}

Hence, for a given block $J = 1, \dots, m$ we need to solve the following two problems:
\begin{itemize}
\item[I.]  For each $j = 1, \dots n_J$, compute a minimizer for the function 
\begin{multline*}
\mathbb{R} \ni \hat x \to \block{Q}{J}(\block{x}{J}, \dots,
\block{x}{J}_{j-1}, \hat x, \block{x}{J}_{j+1}, \dots, \block{x}{J}_{n_J}) \\+
\lambda \block{\Phi}{J}(\block{x}{J}, \dots,
\block{x}{J}_{j-1}, \hat x, \block{x}{J}_{j+1}, \dots, \block{x}{J}_{n_J}).
\end{multline*}

\item[II.] Compute a descent direction at zero for (\ref{eq:inner}).
\end{itemize}

\paragraph{Regarding I} Writing out the equation we see that in the $j$'th iteration we need to find the minimizer of the function $\omega:\mathbb{R} \to \mathbb{R}$ given by
\begin{equation}\label{eq_quadratic_app_coord}
\omega(\hat x) \deff c \hat x + \frac{1}{2}h \hat x^2 +
\gamma\sqrt{\hat x^2 + r} + \xi\abs{\hat x}
\end{equation}
with $c = \block{g}{J}_j + \sum_{i \neq j} (H_{JJ})_{ji}x_i$, $\gamma = \lambda(1-\alpha)\gamma_J$, $\xi = \lambda\alpha\block{\xi}{J}_j$, $r= \sum_{i\neq j} x_i^2$, and where $h$ is the $j$'th diagonal of the Hessian block $H_{JJ}$.

By convexity of $f$ we conclude that $h\ge 0$. 
Lemma \ref{thm:mini1} below deals with the case $h>0$. 
Since the quadratic approximation $Q$ is bounded below the case $h=0$ implies that $c=0$, hence for $h=0$ we have $\hat x = 0$.

\begin{thm_lemma}\label{thm:mini1}
If $h > 0$ then the minimizer $\hat x$ of $\omega$ is given as follows:
\begin{itemize}
\item[a.] If $r = 0$ or $\gamma = 0$ then
\[\hat x = 
\begin{cases}
\frac{\xi + \gamma - c}{h} & \text{if } c > \xi + \gamma \\
0 & \text{if } \abs{c} \leq \xi + \gamma \\
\frac{-\xi - \gamma - c}{h} & \text{if } c < -\xi - \gamma
\end{cases}
\]
\item[b.] If $r > 0, \gamma > 0$ then $\hat x = 0$ if $\abs{c} \le \xi$ and otherwise the solution to
\[
c + \sgn(\xi-c)\xi + h\hat x + \gamma \frac{\hat x}{\sqrt{\hat x^2+r}} = 0.
\]
\end{itemize}
\end{thm_lemma}
\begin{proof}
Simple calculations will show the results.
\end{proof}

For case (b) in the above lemma we solve the equation by applying a standard
root finding method.

\paragraph{Regarding II}
For a convex function $f:\mathbb{R}^n\to\mathbb{R}$ and a point $x\in\mathbb{R}^n$, the vector
\[
\Delta = - \underset{\hat x \in \partial f(x)}{\arg\min}\,\norm{\hat x}_2
 \]
is a descent direction at $x$ provided $f$ is not optimal at $x$, see \cite{bertsekas2003convex}, Section 8.4.
We may use this fact to compute a descent direction at zero for the function (\ref{eq:inner}). 
By Proposition \ref{thm:kprop} it follows that $\Delta \in \mathbb{R}^n$ defined by
\[
\Delta_i \deff -\begin{cases}
0 & \abs{\block{g}{J}_i} \le \lambda\alpha\block{\xi}{J}_i \\
\block{g}{J}_i - \lambda\alpha\block{\xi}{J}_i\sgn(\block{g}{J}_i) &
\text{otherwise}
\end{cases}
\]
is a descent direction at zero for the function (\ref{eq:inner}).

\begin{algorithm}
\caption{Inner loop. Compute the minimizer of (\ref{eq:inner}) by a modified coordinate descent scheme.}
\begin{algorithmic} \label{algo:inner}
\REPEAT
\STATE Choose next parameter index $j$ according to the cyclic rule.
\STATE Compute
\begin{multline*}
\block{x}{J}_j = \argmin{\hat x\in\mathbb{R}} \,
\block{Q}{J}(\block{x}{J}_1, \dots, \block{x}{J}_{j-1}, \hat x,
\block{x}{J}_{j+1}, \dots, \block{x}{J}_{n_J}) \\ +  \lambda
\block{\Phi}{J}(\block{x}{J}, \dots, \block{x}{J}_{j-1}, \hat x,
\block{x}{J}_{j+1}, \dots, \block{x}{J}_{n_J})
\end{multline*}
\IF{$\norm{\block{x}{J}}_2 < \epsilon$ and $\block{Q}{J}(\block{x}{J}) + \lambda
\block{\Phi}{J}(\block{x}{J}) \ge 0$}
\STATE Compute a descent direction, $\Delta$, at zero for (\ref{eq:inner}).
\STATE Use line search to find $t$ such that $\block{Q}{J}(t\Delta) + \lambda
\block{\Phi}{J}(t \Delta) < 0$.
\STATE Let $\block{x}{J} = t\Delta$
\ENDIF
\UNTIL{stopping condition is met.}
\end{algorithmic}
\end{algorithm}

\subsection{Hessian upper bound optimization} \label{sec:ubound}

In this section we present a way of reducing the number of blocks for which the block gradient needs to be computed. 
The aim is to reduce the computational costs of the algorithm. 

In the middle loop, Algorithm \ref{algo:middle}, the block gradient (\ref{eq:blockgradient}) is computed for all $m$ blocks. 
To determine if a block is zero it is, in fact, sufficient to compute an upper bound on the block gradient. 
Since the gradient, $q$, is already computed we focus on the term involving the Hessian.
That is, for $J=1, \dots, m$, we compute a $b_J\in \mathbb{R}$ such that 
\[
\abs{\block{\left[H(x-\beta)\right]}{J}} \preceq b_JD_{n_J}
\]
where $D_n \deff (1,1, \dots, 1) \in \mathbb{R}^n$. 
We define
\[t_J \deff \sup \set{x \geq 0}{\sqrt{K_J(\lambda\alpha\block{\xi}{J},
\abs{\block{q}{J}} + x D_{n_J})} \leq
\lambda(1-\alpha)\gamma_J} 
\]
when $\sqrt{K_J(\lambda\alpha\block{\xi}{J}, \abs{\block{q}{J}})} \le \lambda(1-\alpha)\gamma_J$ and otherwise let $t_J = 0$. 
When $b_J < t_J$ it follows by Lemma \ref{thm:klemma} that
\begin{align*}
K_J(\lambda\alpha\block{\xi}{J}, \block{g}{J}) &=
K_J(\lambda\alpha\block{\xi}{J}, \abs{\block{g}{J}}) \\
&\le K_J(\lambda\alpha\block{\xi}{J}, \abs{\block{q}{J}} + b_JD_{n_J}) \\
&\le \lambda^2(1-\alpha)^2\gamma_J^2
\end{align*}
and by Proposition \ref{thm:kprop} this implies that the block $J$ is zero. 
The above considerations lead us to Algorithm \ref{aglo:modmiddle}.
Note that it is possible to compute the $t_J$'s by using the fact that the function
\[
 \mathbb{R} \ni x \to K_J(\lambda\alpha\block{\xi}{J}, \abs{\block{q}{J}} + x
D_{n_J})
\]
is monotone and piecewise quadratic. 

\begin{algorithm}
\caption{Middle loop with Hessian bound optimization.}
\begin{algorithmic} \label{aglo:modmiddle}
\REPEAT
\STATE Choose next block index $J$ according to the cyclic rule.
\STATE Compute upper bound $b_J$.
\IF{$b_J < t_j$}
\STATE Let $\block{x}{J} = 0$.
\ELSE
\STATE Compute $\block{g}{J}$ and compute new $\block{x}{J}$ (see Algorithm
\ref{algo:middle}).
\ENDIF
\UNTIL{stopping condition is met.}
\end{algorithmic}
\end{algorithm}

In Algorithm \ref{aglo:modmiddle} it is only necessary to compute the block gradient for those blocks where $\block{x}{J} \neq 0$ or when $b_j < t_J$.  
This is only beneficial if we can efficiently compute a sufficiently good bound $b_J$. 
For a broad class of loss functions this can be done using the Cauchy-Schwarz inequality.
 
To assess the performance of the Hessian bound scheme we used our
multinomial sparse group lasso implementation with and without bound optimization (and with $\alpha = 0.5$).
Table \ref{tab:timehb} lists the ratio of the run-time without using
bound optimization to the run-time with bound optimization, on the three
different data sets.  
The Hessian bound scheme decreases the run-time for the multinomial loss function, and the ratio increases with the number of blocks $m$ in the data set.
The same trend can be seen for other values of $\alpha$.

\begin{table} 
\begin{center}
\begin{tabular}{ l r r r}
 Data set & $n \ \ $ & $m \ $ & Ratio \\
\hline
Cancer &  3.9k &217 & 1.14  \\
Amazon & 500k & 10k &1.76 \\
Muscle & 220k & 22k &2.47 \\
\end{tabular} 
\end{center}
\caption{Timing the Hessian bound optimization scheme.}
\label{tab:timehb}
\end{table}

\section{Software} \label{sec:soft}

We provide two software solutions in relation to the current paper.
An R package, \verb¡msgl¡, with a relatively simple interface to our
multinomial and logistic sparse group lasso regression routines. 
In addition, a \verb!C++! template library, \verb¡sgl¡, is provided. 
The \verb¡sgl¡ template library gives access to the generic sparse group lasso routines.
The R package relies on this library. 
The \verb¡sgl¡ template library relies on several external libraries. 
We use the Armadillo \verb!C++! library \cite{nicta_4314} as our primary linear algebra engine.
Armadillo is a \verb!C++! template library using expression template techniques to optimize the performance of matrix expressions.
Furthermore we utilize several Boost libraries \cite{BoostSite}.
Boost is a collection of free peer-reviewed \verb¡C++¡ libraries, many of which are template libraries.
For an introduction to these libraries see for example \cite{demming2010introduction}. 
Use of multiple processors for cross validation and subsampling is supported through OpenMP \cite{openMpSite}. 

The \verb!msgl! R package is available from \verb!CRAN!.
The \verb!sgl! library is available upon request.

\subsection{Run-time performance}

Table  \ref{tab:runtime} lists run-times of the current multinomial sparse group lasso implementation for three real data examples. 
For comparison, the \verb!glmnet! uses 5.2s, 8.3s and 137.0s, respectively, to fit the lasso path for the three data sets in Table \ref{tab:runtime}.
The \verb!glmnet! is a fast implementation of the coordinate descent algorithm for fitting generalized linear models with the lasso penalty or the elastic net penalty \cite{friedman2010regularization}.
The \verb!glmnet! cannot be used to fit models with group lasso or sparse group lasso penalty.

\begin{table}
\begin{center}
\begin{tabular}{lrrrrrr}
\multirow{2}{*}{Data set} & \multirow{2}{*}{$n \ \ $} &
\multirow{2}{*}{$m \ $} & \multirow{2}{*}{Lasso} & \multicolumn{2}{c}{Sparse group lasso} & \multirow{2}{*}{Group lasso} \\
& & & & $\alpha = 0.75$ & $\alpha=0.25$ & \\ 
\hline
Cancer & 3.9k & 217 & 5.9s & 4.8s & 6.3s & 6.0s \\
Muscle & 220k &22k &25.0s & 25.8s & 37.7s & 36.7s \\
Amazon & 500k & 10k & 331.6s& 246.7s & 480.4s &  285.1s \\
\end{tabular}
\end{center}
\caption{Times for computing the multinomial sparse group lasso regression solutions for a lambda sequence of length 100, on a 2.20 GHz Intel Core i7 processor (using one thread).
In all cases the sequence runs from $\lambda_{\max}$ to 0.002.
The number of samples in the data sets Cancer, Muscle and Amazon are respectively 162, 107 and 1500.
See also Table \ref{tab:datasets} and the discussions in Sections \ref{sec:mir}, \ref{sec:mus} and \ref{sec:ama} respectively.} \label{tab:runtime}
\end{table}

\section{Conclusion}
We developed an algorithm for solving the sparse group lasso optimization problem with a
general convex loss function. Furthermore, convergence of the algorithm was established in
a general framework. This framework includes the sparse group lasso penalized negative-log-likelihood for the multinomial distribution, which is of primary interest for multiclass
classification problems.

We implemented the algorithm as a \verb!C++! template library. An R package is available for the
multinomial and the logistic regression loss functions. We presented applications to
multiclass classification problems using three real data examples. The multinomial group
lasso solution achieved optimal performance in all three examples in terms of estimated
expected misclassification error. In one example some sparse group lasso solutions
achieved comparable performance based on fewer features. If there is cost associated with
the acquisition of each feature, the objective would be to minimize the cost while optimizing
the classification performance. In general, the sparse group lasso solutions provide sparser
solutions than the group lasso. These sparser solutions can be of interest for model
selection purposes and for interpretation of the model.

\appendix
\section{Block coordinate descent methods} \label{app:bcd}
In this section we review the theoretical basis of the optimization methods that we apply in
the sparse group lasso algorithm. We use three slightly different methods: a coordinate
gradient descent, a block coordinate descent and a modified block coordinate descent.

We are interested in unconstrained optimization problems on $\mathbb{R}^n$ where the coordinates are naturally divided into $m\in \mathbb{N}$ blocks with dimensions $n_i \in \mathbb{N}$ for $i=1,\dots, m$.
We decompose the search space 
\[
\mathbb{R}^{n} = \mathbb{R}^{n_1}\times\dots\times\mathbb{R}^{n_m}
\]
and denote by $P_i$ the orthogonal projection onto the $i$'th block.
For a vector $x \in \mathbb{R}^n$ we write $x = (\block{x}{1},\dots,\block{x}{m}) $ where $\block{x}{1} \in \mathbb{R}^{n_1},\dots, \block{x}{m} \in \mathbb{R}^{n_m}$. 
For $i = 1,\dots, m$ we call $\block{x}{i}$ the $i$'th \emph{block} of $x$.
We assume that the objective function $F:\mathbb{R}^n \to \mathbb{R}$ is bounded below and of the form 
\[
F(x) = f(x)+\sum_{i=1}^m h_i(\block{x}{i})
\]
where $f:\mathbb{R}^n \to \mathbb{R}$ is convex and each $h_i:\mathbb{R}^{n_i} \to \mathbb{R}$, for $i=1, \dots, m$ are convex.
Furthermore, we assume that for any $i=1,\dots,m$ and any $x_0=(\block{x_0}{1}, \dots, \block{x_0}{m})$ the function 
\[
\mathbb{R}^{n_i}\ni\hat x \to F(\block{x_0}{1}, \dots, \block{x_0}{i-1}, \hat x, \block{x_0}{i+1}, \dots, \block{x_0}{m})
\]
is hemivariate. A function is said to be \emph{hemivariate} if it is not constant on any line segment of its domain.

\subsection{Coordinate gradient descent}

\begin{algorithm} 
\caption{Coordinate gradient descent scheme.}
\begin{algorithmic} \label{algo_newton}
\REPEAT
\STATE Compute quadratic approximation $Q$ of $f$ around the current point $x$.
\STATE Compute search direction 
\[
x^{new} = \argmin{\hat x\in\mathbb{R}^n} Q(\hat x) + \sum_{i=1}^m h_i\left(\block{\hat x}{i}\right).                                
\]
\STATE Let $\Delta = x-x^{new}$ and compute step size $t$ using the Armijo rule and let $x \gets x +
t\Delta$.
\UNTIL{stopping condition is met.}
\end{algorithmic}
\end{algorithm}

\begin{algorithm} 
\caption{Armijo rule.}
\begin{algorithmic} \label{algo_armijo}
\REQUIRE $a  \in (0, 0.5)$ and $b \in (0,1)$
\STATE Let $\delta = \nabla f(x)^T\Delta + \sum_{i=1}^m\left(
h_i(x_i+\Delta_i)-h_i(x_i)\right)$.
\WHILE{$F(x + t\Delta) > F(x) + ta\delta$}
\STATE $t \gets bt.$
\ENDWHILE
\end{algorithmic}
\end{algorithm}

For this scheme we make the additional assumption that $f$ is twice continuously differentiable everywhere.
The scheme is outlined in Algorithm \ref{algo_newton}, where the step size is chosen by the Armijo rule outlined in Algorithm \ref{algo_armijo}. 
Theorem 1e in \cite{tseng2009coordinate} implies the following:
 
\begin{thm_cor}
If $f$ is twice continuously differentiable then every cluster point of the sequence $\{x_k\}_{k\in\mathbb{N}}$ generated by Algorithm \ref{algo_newton} is a minimizer of $F$.
\end{thm_cor}

\subsection{Block coordinate descent}
\begin{algorithm}[h!]
\caption{Block coordinate descent.}
\begin{algorithmic} \label{algo_bcd}
\REPEAT
\STATE Choose next block index $i$ according to the cyclic rule.
\STATE $\block{x}{i} \gets \argmin{\hat x\in \mathbb{R}^{n_i}} F(\hat x \oplus
P_i^\perp x).$
\UNTIL{some stopping condition is met.}
\end{algorithmic}
\end{algorithm}

The block coordinate descent scheme is outlined in Algorithm \ref{algo_bcd}.
By Corollary \ref{thm_bcd} below the block coordinate descent method converges to a \emph{coordinatewise minimum}.

\begin{thm_def}\label{thm:cwmp}
A point $p \in \mathbb{R}^n$ is said to be a coordinatewise minimizer of $F$
if for each block $i = 1, \dots, m$ it
holds that 
\[
 F(p + (0, \dots, 0, d_i, 0, \dots, 0)) \ge F(p)  \text{ for all } d_i \in
\mathbb{R}^{n_i}.
\]
\end{thm_def}

If $f$ is differentiable then by Lemma \ref{thm:sep} the block coordinate descent method converges to a minimizer.
Lemma \ref{thm:sep} below is a simple consequence of the separability of $F$.

\begin{thm_lemma}\label{thm:sep}
Let $p\in\mathbb{R}^n$ be a coordinatewise minimizer of $F$. 
If $f$ is differentiable at $p$ then $p$ is a stationary point of $F$.
\end{thm_lemma}

Proposition 5.1 in \cite{tseng2001convergence} implies the following:

\begin{thm_cor} \label{thm_bcd}
For the sequence $\{x_k\}_{k\in\mathbb{N}}$ generated by the block
coordinate descent algorithm (Algorithm \ref{algo_bcd}) it holds that every
cluster point of $\{x_k\}_{k\in\mathbb{N}}$ is a coordinatewise minimizer of $F$.
\end{thm_cor}

\section{Modified block coordinate descent} \label{app:modbcd}

\begin{algorithm}
\caption{Modified coordinate descent loop.}
\begin{algorithmic} \label{mod_bcd}
\REPEAT
\STATE Let $i \gets i + 1 \mod m.$
\STATE $\block{x}{i} \gets \argmin{\hat x\in \mathbb{R}^{n_i}} F(\hat x \oplus
P_i^\perp x).$
\IF{$\norm{x-p}_2 < \epsilon$ and $F(x) \ge F(p)$}
\STATE Compute descent direction $\Delta$ at $p$ for $F$.
\STATE Use line search to find $t$ such that $F(p+t\Delta) < F(p)$.
\STATE Let $\block{x}{i} \gets p + t\Delta$.
\ENDIF
\UNTIL{stopping condition is met.}
\end{algorithmic}
\end{algorithm}

For this last scheme we make the additional assumption that $f$ is twice continuously differentiable everywhere except at a given non-optimal point $p\in \mathbb{R}^n$.
In this case the block coordinate descent method is no longer guaranteed to be globally convergent, as it may get stuck at $p$.
One immediate solution to this is to compute a descent direction at $p$, then use a line search to find a starting point $x_0$ with $F(x_0) < F(p)$.
Since $f$ is differentiable on the sublevel set $\set{x\in\mathbb{R}^n}{F(x) < F(p)}$ it follows by the results above that the cluster points of the generated sequence are stationary points of $F$. 
This procedure is not efficient since it discards a carefully chosen starting point.
We apply the modified coordinate descent loop, outlined in Algorithm \ref{mod_bcd}, instead.

\begin{thm_lemma} \label{thm:modbcd}
Assume that $f$ is differentiable everywhere except at $p\in\mathbb{R}^n$, and that $F$ is not optimal at $p$.
Then for any $\epsilon > 0$ the cluster points of the sequence $\{x^k\}_{k\in\mathbb{N}}$ generated by Algorithm \ref{mod_bcd} are minimizers of $F$.
\end{thm_lemma}
\begin{proof}
Let $z$ be a cluster point of $\{x^k\}$.
By Corollary \ref{thm_bcd}, $z$ is a coordinatewise minimizer of $F$.
Then Lemma \ref{thm:sep} implies that $z$ is either $p$ or a stationary point of $F$.
We shall show by contradiction that $p$ is not a cluster point of $\{x^k\}_{k\in\mathbb{N}}$, thus assume otherwise.
The sequence $\{F(x^k)\}_{k\in\mathbb{N}}$ is decreasing;
hence, if we can find a $k'\in\mathbb{N}$ such that $F(x^{k'}) < F(p)$ we reach a contradiction (since this would conflict with the continuity of $F$).
Choose $k'$ such that $\norm{x^{k'}-p}_2< \epsilon$. 
Since we may assume that $F(x^{k'}) \ge F(p)$ it follows by the definition of Algorithm \ref{mod_bcd} that $F(x^{k'+1}) < F(p)$.  
\end{proof}

\section{Proof of Proposition \ref{thm:kprop}} \label{sec:proofkprop}
\itemref{a} Straightforward.

\itemref{b} If $\norm{\kappa(v,z)}_2 \le a$ then $-\kappa(v,z) \in aB^n$ hence $0\in X$.
For the other implication simply choose $y_0\in Y$ such that $-y_0 \in aB^n$ and note that $\norm{\kappa(v,z)}_2 \le \norm{y_0}_2 \le a$. 

\itemref{c} Assume $\norm{\kappa(v,z)}_2 > a$, and let $x^* = (1 - a/\norm{\kappa(v,z)}_2)\kappa(v,z)$.
Then $x^* \in X$ and $\norm{x^*}_2 =  \norm{\kappa(v,z)}_2 - a$.
The point $x^*$ is in fact a minimizer.
To see this let $x'\in X$, that is we have 
\[
x' = z + as+\diag(v)t 
\]
for some $s \in B^n$ and $t\in T_n$.
It follows, by the triangle inequality and \itemref{a}, that
\[
\norm{x'}_2 + a \ge \norm{x' - a s}_2 = \norm{z+\diag(v)t}_2 \ge \norm{\kappa(v,z)}_2. 
\]
So $\norm{x'}_2 \ge \norm{\kappa(v,z)}_2-a = \norm{x^*}_2$ and since $X$ is convex and $x\to \norm{x}_2$ is strictly convex the found minimizer $x^*$ is the unique minimizer.

\bibliographystyle{elsarticle-num}
\bibliography{cite}

\end{document}